\documentclass[a4paper,fleqn]{cas-dc}
\usepackage[authoryear]{natbib}
\def\tsc#1{\csdef{#1}{\textsc{\lowercase{#1}}\xspace}}
\tsc{WGM}
\tsc{QE}
\tsc{EP}
\tsc{PMS}
\tsc{BEC}
\tsc{DE}

\usepackage{amsmath} 

\usepackage{graphicx}
\usepackage{soul,color}
\usepackage{xcolor}

\usepackage{hyperref}
\usepackage{stfloats}
\begin{document}

\let\WriteBookmarks\relax
\def\floatpagepagefraction{1}
\def\textpagefraction{.001}

\shorttitle{Collaborative Human-Robot Surgery for Mandibular Angle Split Osteotomy}

\shortauthors{Z.Han etc al.}

\title[mode = title]{
Collaborative Human-Robot Surgery for Mandibular Angle Split Osteotomy: Optical Tracking based Approach
}

\author[1]{Zhe Han}
\author[2]{Huanyu Tian}
\author[3]{Tom Vercauteren}
\author[4]{Da Liu}
\author[2]{Changsheng Li}
\ead{lics@bit.edu.cn}
\author[1,2]{Xingguang Duan}
\fnmark[*]
\ead{duanstar@bit.edu.cn}

\affiliation[1]{organization={School of Medical Technology and the Key Laboratory of Biomimetic Robots and Systems,
        Beijing Institute of Technology},
    city={Beijng},
    postcode={100081}, 
    country={China}}
\affiliation[2]{organization={School of Mechatronical Engineering and  the Key Laboratory of Biomimetic Robots and Systems , Beijing Institute of Technology},
    city={Beijng},
    postcode={100081}, 
    country={China}}
\affiliation[3]{organization={School of Biomedical Engineering \& Imaging Sciences, King's College London},
    city={London},
    postcode={SE1 7EU}, 
    country={UK}}
\affiliation[4]{organization={Remebot Company},
    city={Beijng},
    postcode={100083}, 
    country={China}}

\cortext[cor1]{Corresponding author: Changsheng Li and Xingguang Duan}


\begin{abstract}
Mandibular Angle Split Osteotomy (MASO) is a significant procedure in oral and maxillofacial surgery.
Despite advances in technique and instrumentation, its success still relies heavily on the surgeon's experience. 
In this work,
a human-robot collaborative 
system is proposed to perform MASO according to a preoperative plan and under guidance of a surgeon.
A task decomposition methodology is used to divide the collaborative surgical procedure into three subtasks:
1) positional control and 2) orientation control, both led by the robot for precise alignment; and 3) force-control, managed by surgeon to ensure safety.
Additionally, to achieve patient tracking without the need for a skull clamp, an optical tracking system (OTS) is utilized.
Movement of the patient mandibular is measured with an optical-based tracker mounted on a dental occlusal splint.
A registration method and Robot-OTS calibration method are introduced to achieve reliable navigation within our framework.
The experiments of drilling were conducted on the realistic phantom model, which demonstrated that the average error between the planned and actual drilling points is 1.85mm.


\end{abstract}

\maketitle
\section{Introduction}
The demand for oral and maxillofacial surgery
is steadily growing, both for treating diseases and for aesthetic purposes~\cite{article}. Mandibular angle spilt osteotomy (MASO) is one of the most crucial procedures in craniofacial plastic surgery and oral and maxillofacial surgery.
It involves fundamental operations such as drilling and cutting (osteotomy).
In traditional MASO, surgeons are required to perform the procedure through a narrowed intraoral incision \cite{book}.
Limited visibility and space make it challenging to precisely follow the preoperatively planned path.
Furthermore, the whole procedure can extend over 8 hours \cite{article2}, making it particularly difficult for surgeons to manually operate on bones with handheld tools. These factors exert considerable physical strain on surgeons, negatively affecting the safety and accuracy of the surgery.

Navigation systems have been developed to help surgeons identify the pose of the mandibular and guide the surgery~\cite{balasundaram2012recent}.
In the preoperative planning phase, creating a 3D model
from medical imaging (CT and MRI) is crucial.
This allows surgeons to plan an operative path on the 3D model with great precision and safety.
Intraoperatively, a tracking device is needed to track the surgical tool and transform it into the corresponding model space via a patient registration procedure.
Consequently, the spatial relationship between the surgical tool and the surgical area is established and visualized. This integration ensures that the preoperative planning path is accurately aligned with the intraoperative operational path, enhancing the efficacy and safety of the surgery. 
Although navigation systems have been widely applied in various medical fields, such as dental surgery \cite{CASAP2011512,articlebou,6716056,8985248}, thoracic-abdominal puncture \cite{zheng2020novel}, and craniotomy~\cite{10266731}, its application in MASO is still quite limited.
\cite{shi2017study} implemented augmented reality (AR) technology for the registration of the MASO. In this method, surgeons make the occlusal splint according to a dental cast to fix the marker, which can be recognized by the ARToolKit. When the marker, occlusal splint, and the mandible image of the patient are integrated, the operation plan can be overlaid on the rapid prototyping model of the mandible as soon as the ARToolKit recognized the marker. 
\cite{han2021robot} employed an optical tracking system for image navigation in orthognathic surgery. A registration body was involved in this paper to align preoperative data with CT and physical image spaces. However, this required manual acquisition of sphere positions on the registration body, a process prone to errors. Additionally, the study assessed the efficiency and precision of integrating a robot arm with intraoperative image-guided navigation in orthognathic surgery, highlighting its significant potential and accuracy.

\begin{figure}[t]
	\centering
	\includegraphics[width=\columnwidth]{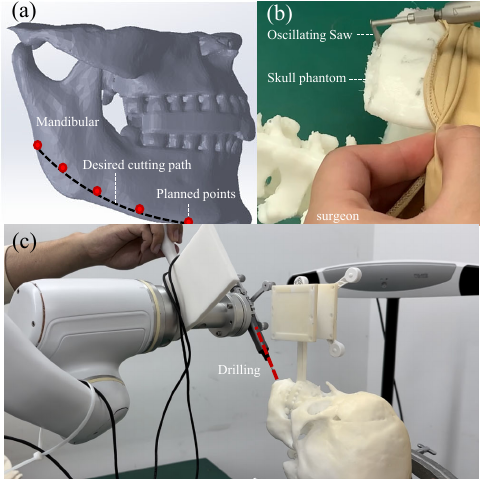}
     \vspace{-10pt}
	  \caption{Path planning and osteotomy procedure for MASO. (a) Schematic of surgical path planning for MASO. Points are preoperatively planned by surgeons, and the cutting curve is generated by the medical imaging software based on the planned drilling points.
   (b) Schematic diagram of a surgeon performing a freehand osteotomy on a skull phantom. (c) The surgeon conducts the drilling operation using a robot-assisted system.}
	  \label{figintro}
    \vspace{-10pt}
\end{figure}

As with navigation systems, robot-assisted surgery is not yet commonly used in MASO.
The motivation for introducing robots into the procedure stems from the challenges surgeons face during prolonged operations, such as the cutting illustrated in Fig.~\ref{figintro}(b).
Both of which can significantly drain a surgeon's physical stamina and mental focus \cite{10081301}.
Surgeon fatigue creates additional risks for the patient, particularly due to the potential loss of accuracy~\cite{10130201}.
Furthermore, prolonged operation times due to surgeon fatigue can lead to more severe complications and inferior surgical outcomes \cite{6678696}.
Conventional robot-assisted orthopedic surgery typically utilizes the robot as a static fixture to hold the tool, which is more similar to a stereotactic device \cite{10228084,8833971}.
In these scenarios, robots are limited to supporting simple actions and are incapable of executing different sub-tasks simultaneously.
In MASO, an interesting feature would be for
the robot to continuously align with the target \cite{9005387}. 
However, it is crucial that this task is not allowed to lead the robot to autonomously make contact with the target, due to significant ethical concerns and patient safety issues \cite{9501975}.
To mitigate these risks,
the approach should
incorporate surgeon supervision, ensuring both precision and safety in the procedure.
\cite{articlezhou} proposed a passive arm and AR feedback to guide surgeons conducting oral and maxillofacial surgery which received good results both in accuracy and ease of use.
However, AR-based systems do not actually offer motion assistance or direct guidance. In essence, the tracking is actually performed by the surgeon.

From the above literature, several challenges currently faced by MASO surgeries can be summarized: 1) In traditional MASO surgeries, surgeons manually bear the weight of the drill and need to expend significant physical effort to perform osteotomy, which is very laborious; 2) There are currently AR-based passive arm systems that help surgeons visualize the osteotomy path, but since the passive arm does not provide active positioning, the positioning accuracy entirely depends on the surgeon's operational precision, failing to leverage the robot's advantages in positioning; 3) Some optical-based MASO robotic systems only perform a one-time patient registration and hand-eye calibration preoperatively, unable to adapt to intraoperative displacements. To enhance the autonomy level of robot-assisted MASO surgery, we employ a human-robot collaborative operating system and utilize an occlusal splint to fix the optical-based tracker to the patient, thus allowing real-time displacement information during surgery to be fed back to the robotic operating system through the optical tracking system (OTS). Surgeons and robotic manipulator play different roles in the drilling task: surgeons lead the feeding/retreating direction and receive the tactile feedback of manual drilling during the process, enabling surgeons to intuitively feel the changes in force during the drilling process; the robot leverages its advantage in positioning accuracy. At the same time, the robot-assisted operating system can solve the problem of human long-term weight bearing and avoid surgery precision reduction caused by excessive fatigue, trembling, etc.

More specifically, in this work, a new robot-assisted system with optical-based navigation and human-robot collaborative operation for MASO is presented to address the limitations on visual and manual capability of surgeons \cite{ma2019development}. Firstly, in line with clinical requirements, a complete robot-assisted MASO operation procedure is outlined.
Surgeons plan the drilling points based on the preoperative 3D model, as shown in Fig. \ref{figintro} (a). 
Five planned drilling points are selected, and a smooth connecting line through these points determines the osteotomy path.
Secondly, optical-based trackers are appropriately installed on both the patient and the robot.
Patient registration is completed through preoperative-intraoperative registration with trackers installed on the occlusal splint,
and the robotic manipulator is registered via hand-eye calibration. 
Finally, the collaborative drilling task is divided into three sub-tasks: 1) positional control, 2) orientation control, and 3) force control.
The first two sub-tasks require aligning the surgical tool with the planned drilling points using the OTS.
For the third sub-task, surgeon could apply a slight force in the feeding direction to complete the drilling operation and can sense changes in force during the process, thereby controlling the progress of the operation, as illustrated in Fig. \ref{figintro} (c).

The remainder of this paper is organized as follows. Section~\ref{sec:system} describes the system design and the operation pipeline which includes patient registration and hand-eye calibration. In Section~\ref{sec:methods}, the surgical plan and collaborative control method are presented, aimed at assisting surgeons before and during operations. Section~\ref{sec:experiments} reports the experimental results. The analysis of all the experiments is discussed in Section~\ref{sec:discussion}. Section~\ref{sec:conclusion} concludes this paper.

\section{System Design and Operation Pipeline}
\label{sec:system}

This part describes the robot system design and operation pipeline of robot-assisted MASO surgery. It is essential for surgeons to plan drilling points based on the preoperative patient CT model. Through patient registration and hand-eye calibration, these planned drilling points can be transformed from the patient CT model frame to the camera frame, then to the robot frame, enabling precise positioning. In robot-assisted MASO surgeries, robot tracking is driven by patient movement captured by the OTS \cite{doi:10.34133/cbsystems.0063}.

\begin{figure}[t]
	\centering
	\includegraphics[width=\columnwidth]{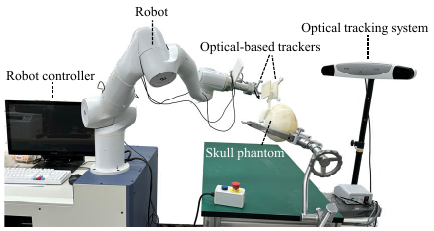}
     \vspace{-10pt}
	  \caption{The schematic diagram of robot-assisted MASO system. It consists of robotic manipulator, optical tracking system and the skull phantom.}
	  \label{fig1}
    \vspace{-10pt}
\end{figure}

\subsection{System Design}

The human-robot collaborative MASO system depicted in Fig.\ref{fig1} primarily encompasses an OTS and a robotic manipulator.

The OTS (Passive Polaris Spectra, NDI, Waterloo, ON, Canada), which acquires the coordinates of optical-based trackers and provides feedback for robotic tracking. The system integrates two optical-based trackers which we refer to as tracker for short. One tracker is installed on an occlusal splint that is fixed to the patient. It is fitted with four non-collinear Radix Lenses, hereafter referred to as “semi-balls”. Noticeably, the Radix Lenses is not a perfect sphere but a shape like a fedora hat with a sphere top.  The other tracker is fitted with four non-collinear NDI Passive Spheres. It is integrated with the surgical tool and mounted at the end of the robotic manipulator.

A seven degrees of freedom robotic manipulator is applied (Diana-7, Agile Company, China), enabling precise control of surgical instrument. The end of the manipulator incorporates two orthogonal force sensors that enable interactions between the human and the robot (HRI) as well as between the environment and the robot (ERI), as discussed in \cite{9361644}. The surgical tool and tracker are fixed to the manipulator's end flange through connectors. All the connectors are made of aluminum alloy, while the handle is crafted from nylon, both to reduce weight and maintain considerable stiffness.

For our experiments, a 3D-printed skull phantom, derived from a patient’s CT scan model of a MASO surgery, was made of PLA material. To support our phantom experiments, the phantom model includes an oral occlusal splint bonded to the teeth of the skull, addressing the necessary tracking issues when the mandible moves.
 
Two main controllers were applied in this system: the navigation processing unit and the robot controller, both based on Ubuntu 18.04. The navigation processing unit was equipped with a high-performance AMD RYZEN 3900XT CPU and an NVIDIA GeForce RTX 2060 GPU, tasked with processing image data. Concurrently, the robot controller, powered by an Intel Core i9-11900K CPU, was responsible for the direct control of the robot. Communication between the two was achieved in real time through a TCP/IP data transmission protocol, ensuring efficient coordination between the robotic manipulator and the navigation system. The navigation software was developed using Python within Visual Studio 2017, while the robot control module was coded in C++.

\subsection{Robot-assisted MASO Operation Pipeline}
\label{sec:operation pipeline}
This section describes the surgical opertion pipeline of robot-assisted MASO. 
The patient needs to wear an occlusal splint prior to undergoing a CT scan. 
Once the CT scan is completed, the surgeon can plan the preoperative surgical path based on the patient's medical images.
To ensure the reliable implementation of the robot-assisted system in this surgery, a drilling-cutting method \cite{yang2009mandibular, park2014burring} is adopted. 
The surgeon needs to select several drilling points as planned points for the cutting path, and commercial medical software will plan the cutting path based on the selected drilling points.
These planned points are manually selected by the surgeon on the medical imaging user interface based on the desired mandibular shape of the patient. 
With the introduction of the robot-assisted system, the coordinates of the drilling points need to be transformed from the image frame (patient CT model frame) to the robot frame, which can be achieved through patient registration and hand-eye calibration. The patient registration, detailed in \ref{sec:patient registration}, aims to complete the transformation of drilling point coordinates from the image frame to the camera frame (OTS frame). 
The hand-eye calibration, discussed in \ref{sec:handeye}, mainly completes the transformation of drilling point coordinates from the camera frame to the robot frame. 
After completing these two steps, the goal of preoperative-intraoperative path registration can be achieved. 
As this paper adopts a frameless surgical approach, the patient's position can be adjusted intraoperatively as needed by the surgeon, and the tracking performance of the system proposed in this paper is introduced in \ref{sec:coopdrillexp}. 
Once the patient's position is stable, the robot achieves precise positioning, allowing the surgeon to guide the manipulator to perform the drilling operation.

\subsection{Patient Registration}
\label{sec:patient registration}
Patient registration is foundational to the navigation accuracy of surgical robots. It provides the positional information of the target and builds the relationship between the frame of the navigation device (camera) and the frame of medical imaging (patient CT model) before surgery. In our scenario, a head CT \cite{creighton2020early} is acquired to obtain the patient's anatomical features and facilitate the formulation of a surgical plan.

\begin{figure}[htb]
	\centering
	\includegraphics[width=\columnwidth]{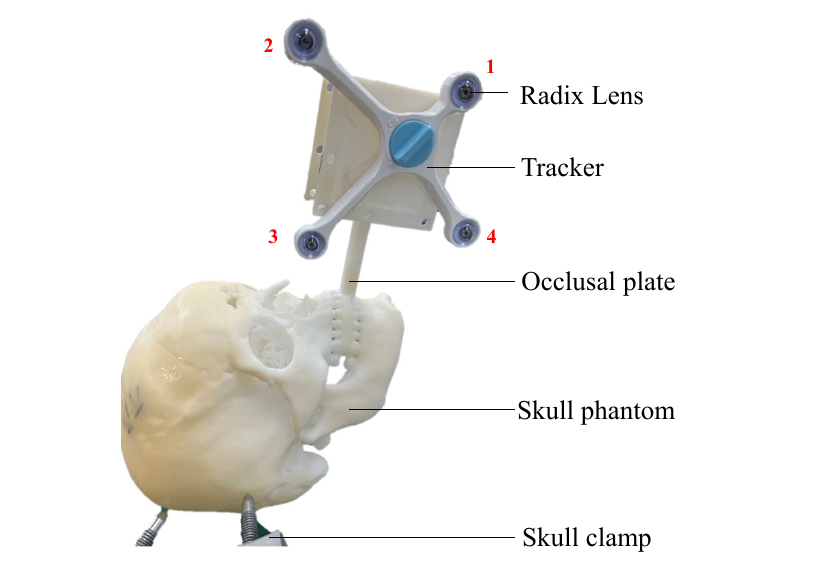}
	  \caption{The optical-based tracker which is outfitted with four non-collinear Radix Lenses. The tracker is fixed on the occlusal splint which is bonded to teeth before CT scan and surgery.}
	  \label{fig2}
\end{figure}

We conduct tracker registration between CT image frame and the camera frame.
For accurate real-time data capture by the OTS that does not interfere with surgical operations, we secure the tracker to the patient using an occlusal splint.
This occlusal splint, depicted in Fig.~\ref{fig2}, is bonded to the teeth and extends outwards to connect with the tracker, maintaining a stable relationship with the mandible.
The design of the occlusal splint is adaptable, altering shape as necessary to fit the specific surgical context.

Before registration, it is crucial to extract features to identify corresponding points on the CT images.
The tracker, also shown in Fig.~\ref{fig2}, includes four non-collinear semi-balls. Since the OTS could precisely identify the centers of these semi-balls, we employ ball-center finding regression in our proposed method to refine the registration process.

The regression process requires a pipeline including point cloud segmentation, estimation of ball centers, and sorting of corresponding points.
Initially, an interactive segmentation method \cite{MALEIKE200972} is employed to isolate the semi-balls on the target, yielding results in STL format.
To generate the point cloud for the semi-balls, we convert the STL files of the semi-balls into a point cloud by sampling the vertex points from a 3D view in medical imaging software where surgeons can fully see all semi-balls.
Once the point cloud is acquired, we can determine both the centers of the balls and their corresponding order.

For semi-ball center finding regression, Gaussian Mixture Models (GMMs) are utilized to cluster point cloud data. This method hypothesizes that the data originates from four distinct Gaussian distributions, each representing a separate cluster. The clustering employs the Expectation-Maximization (EM) algorithm, which iteratively assigns points to the most probable distribution that represents a cluster. 

A key hyper-parameter in this clustering process is the predetermined number of categories to be clustered. In GMMs, the mean values of each Gaussian distribution could serve as the estimated centers of the semi-balls. However, the OTS identifies the centers of whole spheres for four Radix Lens, which is not equal to the mean values of each Gaussian distribution. Equations \eqref{equ1a} and \eqref{equ1b} are used to calculate the centers of the whole spheres. 

%
\begin{subequations}
\begin{equation}
\begin{aligned}
H(\mathbf{x_c}, R) =\sum_{i=1}^{N_{S}}(||\mathbf{x}_i-\mathbf{x_c}||_{2}^{2} - R^{2})^{2}
    \label{equ1a}
    \end{aligned}
\end{equation}
\begin{equation}
\begin{aligned}
\frac{\partial H}{\partial \mathbf{x_c}} = \mathbf{0}, \;\frac{\partial H}{\partial R} = 0
    \label{equ1b}
    \end{aligned}
\end{equation}
\end{subequations}
\begin{figure*}[ht]
\begin{subequations}
\begin{equation}
\begin{bmatrix}
\sum \mathbf{x}_{i,1}^{2}-N_{s}\mathbf{\Bar{x}_1}^{2} &\sum \mathbf{x}_{i,1}\mathbf{x}_{i,2}-N_{s}\mathbf{\Bar{x}_1}\mathbf{\Bar{x}_2} & \sum \mathbf{x}_{i,1}\mathbf{x}_{i,3}-N_{s}\mathbf{\Bar{x}_1}\mathbf{\Bar{x}_3}\\
\sum \mathbf{x}_{i,1}\mathbf{x}_{i,2}-N_{s}\mathbf{\Bar{x}_2}\mathbf{\Bar{x}_1} &\sum \mathbf{x}_{i,2}^{2}-N_{s}\mathbf{\Bar{x}_2}^2 & \sum \mathbf{x}_{i,2}\mathbf{x}_{i,3}-N_{s}\mathbf{\Bar{x}_2}\mathbf{\Bar{x}_3}\\
\sum \mathbf{x}_{i,1}\mathbf{x}_{i,3}-N_{s}\mathbf{\Bar{x}_3}\mathbf{\Bar{x}_1} &\sum \mathbf{x}_{i,2}\mathbf{x}_{i,3}-N_{s}\mathbf{\Bar{x}_2}\mathbf{\Bar{x}_3} & \sum \mathbf{x}_{i,3}^{2}-N_{s}\mathbf{\Bar{x}_3}\mathbf{\Bar{x}_3}\\
\end{bmatrix}\mathbf{x_c} = \mathbf{b}
\label{equ2a}
\end{equation}
\begin{equation}
\mathbf{b} =
\sum \frac{1}{2}\begin{bmatrix}
\mathbf{x}_{i,1}^{3} +\mathbf{x}_{i,1}^{1}\mathbf{x}_{i,2}^{2} + \mathbf{x}_{i,1}^{1}\mathbf{x}_{i,3}^{2}-M_1\mathbf{x}_{i,1}\\
\mathbf{x}_{i,2}^{3} +\mathbf{x}_{i,2}^{1}\mathbf{x}_{i,1}^{2} + \mathbf{x}_{i,2}^{1}\mathbf{x}_{i,3}^{2}-M_1\mathbf{x}_{i,2}\\
\mathbf{x}_{i,3}^{3} +\mathbf{x}_{i,3}^{1}\mathbf{x}_{i,1}^{2} + \mathbf{x}_{i,3}^{1}\mathbf{x}_{i,2}^{2}-M_1\mathbf{x}_{i,3}
\end{bmatrix}
\label{equ2b}
\end{equation}
\end{subequations}
\end{figure*}
where $H$ indicates the loss function of ball-center finding regression. As the problem is convex, the point with partial derivation being zero (shown in  \eqref{equ1b}) is a global optimized point for loss function \eqref{equ1a}. $S$ indicates the subspace of all sampled points on spheres and $\mathbf{x_c} \in \mathcal{R}^{3}$ is a vector including the 3 channels of center point's coordinates. $R$ is the ball's radius, which is an unknown parameter in regression. Besides, $\mathbf{x}_i \in \mathcal{R}^{3}$ represents the i-th point's coordinates which also has 3 channels for representation. 
In our method, RAndom SAmple Consensus (RANSAC) is employed for random sampling of the point cloud to obtain the selected points $\mathbf{x}_i$, which means that we don't use every $\mathbf{x}_i$ in the point cloud (assume there are $N$ points) but sample $N_s$ points($N_s < N$). In our cases, we choose $N_s = 0.2* N$.

To simplify the equations \eqref{equ1a} and \eqref{equ1b}, a regression problem could be reformulated shown in Equations \eqref{equ2a} and \eqref{equ2b}.

$\mathbf{b}$ vector could be represented with $\mathbf{x}_i$ and for shake of brevity, the values in 1st, 2nd, and 3rd channel of the selected points could be represented with $\mathbf{x}_{i,1}$, $\mathbf{x}_{i,2}$, and $\mathbf{x}_{i,3}$, respectively. $\mathbf{\Bar{x}_1}$, $\mathbf{\Bar{x}_2}$, $\mathbf{\Bar{x}_3}$ represent the 1st, 2nd, 3rd channel of $M_{2} = \frac{1}{N_S} \sum_{i=0}^{N_S}\mathbf{x}_i^{T}$, respectively. $M_1 = \frac{1}{N_S} \sum_{i=0}^{N_S}\mathbf{x}_i^{T}\mathbf{x}_i$ and the details are explained in the appendix, which describes the derivation process of Equations \eqref{equ2a} and \eqref{equ2b}.

According to the reformulated representation shown in equations \eqref{equ2a} and \eqref{equ2b}, a $Ax = b$ shaped regression problem could be built and to solve the ball-center finding problem, it's a direct and efficient way to use inverse matrix of $A$.
In our cases, we tried massive tests of the sphere regression. According to the experience, the matrix $A$ never becomes ill-posed or not-full rank. 
Therefore, the solution could be obtained via $x = A^{-1}b$.

The four Radix Lenses are sequentially arranged as depicted in Fig.~\ref{fig2}, where each pair of centers is separated by a distinct distance. 
Upon calculating the centers of the Radix Lenses, we can establish their sequence by selecting any Radix Lens and measuring the distance from its center to the centers of the others, thus establishing their sequential position.
Additionally, to obtain the registration information, another cost function could be given:
\begin{equation}
\begin{aligned}
C_{o}= \sum_{i=1}^{4} ||^{nav}\mathbf{R}_{img} \;^{img}\mathbf{x}_i + ^{nav}\mathbf{t}_{img} - ^{nav}\mathbf{x}_{i}||_{2}^{2}
\label{equ3}
\end{aligned}
\end{equation}
where $^{nav}\mathbf{R}_{img}$ and $^{nav}\mathbf{t}_{img}$ represent the rotation matrix and translate vector from medical imaging frame ($img$) to the navigation device's frame ($nav$), respectively.
Besides, $^{img}\mathbf{x}_i$ denotes positions of the ball-center points in medical imaging frame (CT frame) and the positions of those in navigation device's frame (Camera frame) are denoted with $^{nav}\mathbf{x}_i$. As there are only 4 balls, the iteration variable $i$ could vary from 1 to 4. 

As a result, the output of registration could be solved with the optimization problem:
\begin{equation}
\begin{aligned}
^{nav}\mathbf{R}_{img}^{*}, ^{nav}\mathbf{t}_{img}^{*}= \mathop{argmin}_{^{nav}\mathbf{R}_{img}, ^{nav}\mathbf{t}_{img}}C_{o}
\label{equ4}
\end{aligned}
\end{equation}
where the $^{nav}\mathbf{R}_{img}$ and $^{nav}\mathbf{t}_{img}$ are the values mentioned in equation \eqref{equ3}. To address the minimization problem, singular value decomposition (SVD) can be applied and the details could be found in \cite{sorkine2017least}.

\subsection{Hand-Eye Calibration}
\label{sec:handeye}
\begin{figure}[t]
	\centering
	\includegraphics[width=\columnwidth]{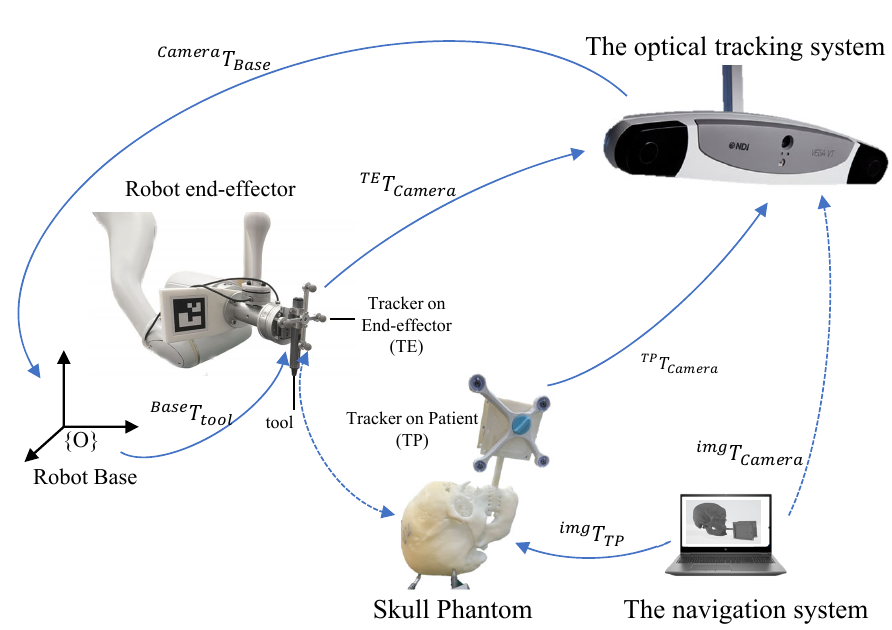}
     \vspace{-10pt}
	  \caption{Illustration of the proposed Robot-assisted system.}
	  \label{figtransformation}
    \vspace{-10pt}
\end{figure}

Hand-eye calibration, which involves obtaining the matrix between the camera frame and the robot frame. In this paper, an eye-to-hand configuration is adopted, as shown in Fig.~\ref{figtransformation}. After completing this step, the preoperative path planned by surgeons can be mapped to the robot frame.
\begin{equation}
    {}^{img}T_{tool} = {}^{img}T_{TP}{}^{TP}T_{Camera}{}^{Camera}T_{Base}{}^{Base}T_{tool}
    \label{equ2}
\end{equation}
where the ${}^{Camera}T_{Base}$ is the result of hand-eye calibration and ${}^{TP}T_{Camera}$ is the positional information of the tracker on patient captured by the OTS, and $TP$ means the tracker on patient;
Both matrices, ${}^{img}T_{TP}$, representing the relationship between the imaging frame and the patient's tracker, and ${}^{Camera}T_{Base}$, are constant. ${}^{Base}T_{tool}$ denotes the relationship between the tracker on robot and the base of the robot, which can be obtained by the robot's kinematics. Therefore, the manipulator can accomplish accurate positioning tasks.

Accurate Robot-OTS calibration is a crucial requirement during operation. 
As shown in Fig.~\ref{figtransformation}, the position of the manipulator and the OTS remains fixed, and the position between the tool and the tracker on the end-effector is also constant. Based on the aforementioned relationship, we can express the relationship between the manipulator and the OTS in two different poses using Equations \eqref{equ6a} and \eqref{equ6b}:
\begin{subequations}
\begin{align}
{}^{tool}{T_{1}}_{Base}{}^{Base}T_{Camera}{}^{Camera}{T_{1}}_{TE} = {}^{tool}T_{TE}
\label{equ6a}\\
{}^{tool}{T_{2}}_{Base}{}^{Base}T_{Camera}{}^{Camera}{T_{2}}_{TE} = {}^{tool}T_{TE}
\label{equ6b}
\end{align}
\end{subequations}
where the ${}^{tool}{T_{1}}_{Base}$ and ${}^{tool}{T_{2}}_{Base}$ mean the first and the second pose of the manipulator, respectively. ${}^{Camera}{T_{1}}_{TE}$ and ${}^{Camera}{T_{2}}_{TE}$ represent corresponding poses of the tracker on the robot end effector captured by the OTS.

By combining Equations \eqref{equ6a} and \eqref{equ6b}, we obtain
the following relationship:
%
\begin{equation}
    AX=XB
\label{equ7b}
\end{equation}
%
where
$A$ means $({}^{tool}{{T_{2}}_{Base}})^{-1}{}^{tool}{T_{1}}_{Base}$, $B$ means 

\noindent ${}^{Camera}{T_{2}}_{TE}({{}^{Camera}{T_{1}}_{TE}})^{-1}$, and $X$ means ${}^{Base}T_{Camera}$. 
Tsai's method \cite{34770} is adopted to solve for the matrix 
${}^{Base}T_{Camera}$.

\section{Surgical Planning and Collaborative Control}
\label{sec:methods}
This section details the surgical planning and collaborative operation aspects in our proposed MASO approach.
For surgical planning, the drilling points along the cutting path are established using commercial imaging systems, and the orientation for cutting is also specified within the system. 
The robot-assisted MASO system in this paper utilizes collaborative operation. More specifically, the drilling operation is divided into three sub-tasks: positional and orientation control, which autonomously aligns the robot's motion via navigation; force control, which executes transparent human-robot interaction in the feeding/retreating direction during surgical drilling.

\subsection{Surgical Planning for Drilling}
In terms of the cutting path, a feasible method to achieve cutting-path planning requires surgeon to select several planned points to determine a specified path. The planned points are selected manually in the surgical design software. After patient registration and hand-eye calibration, 
robots are commanded to drill through the points separately. With the drilled holes, surgeons/robots~\cite{10081301} could cut along the path with a saw. In consequences, the drilling-point task's accuracy indicates the precision and safety of the surgery.

Considering that the bone surface of the mandible can be approximated as a spatial plane \cite{wang2012real, 9907401}, the orientations during co-manipulation are typically stable. To determine the orientation for a given path, represented by five planned points, we introduce a PCA-based algorithm \cite{pascoletti2021stochastic}:
\begin{subequations}
\begin{equation}
    \begin{aligned}
    \mathbf{Cov=svd(X)}=\sigma_{1}\mathbf{v_{1}v_{1}^{H}}+\sigma_{2}\mathbf{v_{2}v_{2}^{H}}+\sigma_{3}\mathbf{v_{3}v_{3}^{H}}
    \end{aligned}
 \label{equ4a}   
\end{equation}
\begin{equation}
 k^{*} = \mathop{\arg\min}\limits_{k}\ \ \sigma_{k}
\label{equ4b}   
\end{equation}
\end{subequations}
where $\mathbf{X}$ denotes the product of matrices for the stacked planned points, defined as $\mathbf{X = X_t X_{t}^{T}}$ and 

$\mathbf{X_t} = \begin{bmatrix}
    x_1 & x_2& \cdots& x_N\\
    y_1 & y_2& \cdots& y_N\\
    z_1 & z_2& \cdots& z_N\\
\end{bmatrix}$, and $N=5$.
Here, svd($\cdot$) represents the Singular Value Decomposition (SVD), utilized to derive the eigenvector \(\mathbf{v}\) and the eigenvalue \(\sigma\). The subscripts in this notation indicate the decomposition in different channels, ranging from 1 to 3. To determine the orientation of the surgical tool, we leverage the eigenvector corresponding to the smallest eigenvalue as the tool's axial vector. The tool's remaining two axes can be arbitrarily defined, provided the robot's configuration permits such flexibility.

\subsection{Collaborative Control in Operation}
To effectively implement the planned path in collaborative MASO, robotic control necessitates closed-loop positioning for enhanced accuracy. Nonetheless, achieving complete autonomy in robotic surgical procedures remains a future objective \cite{doi:10.1126/scirobotics.aam8638}, primarily due to the indispensable involvement of surgeons in managing unexpected occurrences and unpredictable scenarios. This requirement is further reinforced by safety regulations and ethical considerations.
To incorporate surgeon supervision, physical Human-Robot Interaction (pHRI) is applied within the robot control framework, which could facilitate transparent and intuitive guiding interaction.

In our cases, the force control and positional control could be decomposed into the two subspaces: $S = \mathbf{v}_{k^{*}}$ and $S_{\perp} = \{s| s = a\mathbf{v}_{i} + b\mathbf{v}_{j}\; for\; i, j\; in\; (1,2,3)\; and\; i\neq k^{*}, j\neq k^{*}\; a, b \in \mathcal{R}^{1}\}$. In subspace $S$, admittance-based pHRI control \cite{IJRRAdmittanceControl,8685113} is conducted with the following control laws:
\begin{equation}
\dot{ \mathbf{q}}_{\parallel} = \mathbf{J}^{+}\mathbf{T}_{R}\mathbf{A} \mathbf{v}_{k^{*}} \mathbf{v}_{k^{*}}^{T}\mathbf{f}_{HRI}
\label{equ9}   
\end{equation}
where $\mathbf{q}$ denotes the joint position of the robot and the velocities accordingly is denoted with $\dot{ \mathbf{q}}_{\parallel}$. $\mathbf{J}^{+}$ represents the Jacobian matrix's inverse matrix. The Jacobian matrix is derived from robotic chain (from base to surgical tool) and the geometric properties, which is exactly the geometric Jacobian matrix. $\mathbf{T}_{R}$ represents the stacked matrix of the rotation matrices which transfer the wrench from Force sensor's frame to surgical tool frame. Admittance matrix $\mathbf{A}$ represents the ratio of twist to wrench in HRI sensor's frame. $\mathbf{f}_{HRI}$ is the wrench including translate force and torques recorded from 6 dimensional force sensor on the end effector.

In the collaborative control mode, the autonomous part controls the robot to align to the planned drilling points. 
During the surgery, the surgeon continuously adjusts the patient's head position and posture to ease the drilling process. Consequently, the planned drilling points are not fixed and will move in response to changes in the patient's head position and orientation.
To track the target point in subspace $S_{\perp}$, the control law is given as follows:
\begin{equation}
\dot{ \mathbf{q}}_{\perp} = \mathbf{J}^{+}\mathbf{T}_{R}(I - \mathbf{v}_{k^{*}}^{T}\mathbf{v}_{k^{*}})\mathbf{T}_{R}^{T}(\mathbf{K}\mathbf{e}_t + \mathbf{B}\dot{\mathbf{e}_t})
\label{equ10}   
\end{equation}
where $\mathbf{e}_t = \mathbf{x}_t - \mathbf{x}_e$ denotes the distance between target position and actual position. $\dot{\mathbf{e}_t}$ is the derivation of $\mathbf{e}_t$.
Furthermore, the control parameters $\mathbf{K}$ and $\mathbf{B}$ represent the proportional weighting and derivative weighing in the overall controller. As the decomposition subspaces S and $S_{\perp}$ are orthogonal, the two terms $\dot{ \mathbf{q}}_{\perp}$ and $\dot{ \mathbf{q}}_{\parallel}$ are independent.
Therefore, they can be directly added together. 

Also, orientation control is required in our drilling task as the axis of the drilling tool should be aligned with the normal vector to the mandibular surface.
Without loss of generality, the rotation representation of each planned point can be denoted as $\mathbf{R}_{t}$.
The autonomous drilling control is derived based on the target and current tool orientation $\mathbf{R}_{r}$, which is shown as follows:
\begin{equation}
\dot{ \mathbf{q}}_{rot} = \mathbf{J}^{+}\mathbf{T}_{R} v_{r^{*}} v_{r^{*}}^{T}\mathbf{K}_{rot}(\mathbf{R}_{r}^{T}\mathbf{R}_{t})^{\vee}
\label{equ11}  
\end{equation}
where the vector $\dot{ \mathbf{q}}_{rot}$ represents the vector component responsible for the robot's autonomous rotation.
Analogous to $\mathbf{v}_{k^{*}}$ being the projection vector in force space, variable $v_{r^{*}}$ is a projection vector from Cartesian space to orientation space. $\mathbf{K}_{rot}$ denotes a gain matrix from orientation difference to command in Cartesian space (which could be defined as a gain-modified angular velocity).
Additionally, the sign $\vee$ here represents the reformulation that transfers the skew-symmetric representation to a 3D vector.

As a result, we have the collaborative controller:
\begin{equation}
\dot{ \mathbf{q}} = \dot{ \mathbf{q}}_{\perp}+\dot{ \mathbf{q}}_{\parallel} +\dot{ \mathbf{q}}_{rot}
\label{equ6}   
\end{equation}
where the $\dot{ \mathbf{q}}$ are treated as the command to the control box of robot.

\section{Experiments and Results}
\label{sec:experiments}
There are four types of experiments carried out in this paper, including patient registration, hand-eye calibration, target tracking and drilling.
In the first experiment, by aligning the CT scan model information of the semi-balls on the patient-affixed tracker with the corresponding semi-balls information in the camera frame, patient registration can be completed. This ensures that the preoperative planning path corresponds to the actual surgical scenario during the operation.
In the second experiment, the manipulator was introduced by the hand-eye calibration.
The third experiment was developed for the MASO surgery to enable real-time patient tracking and co-manipulation, which further clarifies the surgeon's role during the surgical procedure.
In our fourth and last experiment, the drilling results were used to prove the effectiveness of this system.

\subsection{Patient Registration} 
\begin{figure*}[t]
    \centering
    \includegraphics[width=\textwidth]{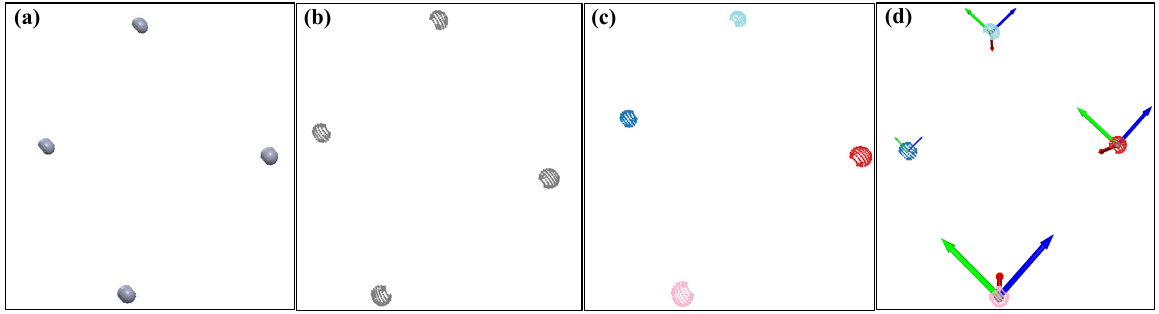}
    \vspace{-10pt}
    \caption{Semi-balls recognition algorithm workflows. (a) 3D
model of semi-balls segmentation result from CT scan model. (b) 
Point cloud model of semi-balls. (c)  Clustering results (each semi-ball point cloud is denoted
with a distinct color). (d) Center coordinates of each semi-ball and the variation in the size of the coordinate system displays the sorting results.}
    \label{BallRegistration}
    \vspace{-10pt}
\end{figure*}

Fig.~\ref{BallRegistration}(a) shows the 3D segmentation results of the semi-balls from CT scan model of the tracker. Fig.~\ref{BallRegistration}(b) shows the point clouds for each semi-ball. 
According to the known configuration of the optical-based tracker, the number of cluster hyper-parameter for the clustering algorithm is set to 4. 
The GMMs result is shown in Fig.~\ref{BallRegistration}(c). The ball-center finding results and the sorting results are shown in Fig.~\ref{BallRegistration}(d).

The distances between the centers of every two balls in the model are used to establish correspondences with the balls in the camera coordinate system. 
In Equation \eqref{equ3}, the four pairs of known values $^{img}\mathbf{x}_{i}$ and $^{nav}\mathbf{x}_{i}$ can be used to calculate $^{nav}\mathbf{R}_{img}$ and $^{nav}\mathbf{t}_{img}$. Using the computed $^{nav}\mathbf{R}_{img}$ and $^{nav}\mathbf{t}_{img}$, the distance between each pair of semi-balls after registration can be calculated. The observed accuracy is 0.32mm, 0.30mm, 0.28mm and 0.22mm, separately.

\subsection{Hand-Eye Calibration}
The pipeline of hand-eye calibration has been explained in Section \ref{sec:handeye}. 
Theoretically, we need to solve Equation~\eqref{equ7b} by collecting three sets of different poses of ${{}^{tool}T_{Base}}$ and ${{}^{Camera}T_{TE}}$.
A rapid calibration method by robot movement is developed in this paper. The manipulator can be manually moved by users randomly. Combined with the MASO surgical scenarios, the user should move the manipulator around the lesion area for uniform sampling, which could ensure high accuracy in navigation around local areas.
In this experiment, 5 different poses were collected.
This could help to reduce the overall calibration time.
Five additional points were collected above the lesion area for testing. By substituting these points into Equation~\eqref{equ7b}, we determined the magnitude of the translational component of $AX-XB$, representing the error in hand-eye calibration. The maximum observed error was 0.65mm.
Based on the experiments, it was found that the time taken for a surgeon to collect all registration points was approximately 90 seconds, which was acceptable during surgery.

\subsection{Evaluations of Tracking under Collaborative Drilling}
\label{sec:coopdrillexp}
\begin{figure*}[htb]
    \centering
    \includegraphics[width=\textwidth]{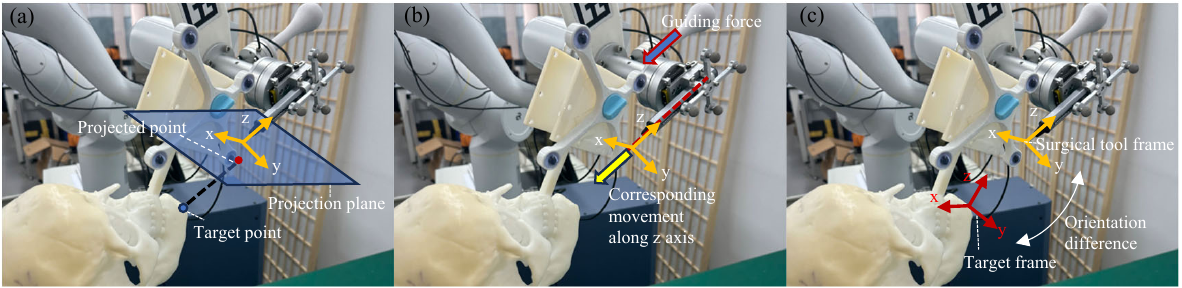}
    \vspace{-10pt}
    \caption{Three sub-tasks during collaborative drilling. (a) Positional control: This sub-task projects the target point onto the plane formed by the x/y axes of surgical tool frame. The positioning error is represented by the 2D distance from the projected point (red point) to the original point of the surgical tool frame (the orange frame) (b) Force control: In this sub-task, an admittance control strategy for HRI transfers the guiding force into the movement along the z-axis of the surgical tool frame. (c) Orientation control: This sub-task is responsible for aligning the orientation of the surgical tool frame with the target frame (red frame).}
    \label{figExperiment}
    \vspace{-10pt}
\end{figure*}

The robot tracking performance was evaluated by moving the skull phantom from an initial position to any location.
To assist the surgeon during the surgery, the robot focused on planar positioning task, ensuring the alignment of the surgical tool with the planned points in the x and y coordinates, shown in Fig. \ref{figExperiment}(a).
Simultaneously, the orientation task focused on aligning the Z axis of the surgical tool with the target frame, as illustrated in Fig. \ref{figExperiment}(c). The feeding direction of the surgical tool is guided and controlled by the surgeon, with the force being recorded by HRI force sensor, shown in Fig.~ \ref{figExperiment}(b).
It is important to note that the use of a head clamp should be avoided during surgery to prevent potential damage to the skull.
The presence of a head clamp in this paper is merely for experimental convenience, reducing the random and unnatural movements caused by the lack of the neck constraint.
After the surgeon drags the robot into the field of view of the OTS,
the related poses between planned point and surgical tool reduced and therefore linearization in robotic control/model, as per equations \eqref{equ9}, \eqref{equ10} and \eqref{equ11}, became effective.

Based on the collaborative operation mode, the evaluation of the tracking performance can be represented using positional error, orientation error, and human-robot interactive force.
Specifically, the positional error referred to the deviation of the surgical tool's axis from the planned point, while the orientation error indicates the angle value in angle-axis representation between the desired pose and the actual pose. The human-robot interactive force represents the force received by the force sensor in the feeding/retreating direction.

Five experiments were conducted to evaluate the tracking performance of the robot.
To assess the system's performance in the presence of disturbances that are primarily caused by movements of the skull clamp and potential crosstalk among positional, orientation, and force control. These disturbances were randomly introduced during the tracking procedures, with the associated scenarios depicted in Fig. \ref{figExperiment}.

\begin{figure*}[htbp]
    \centering
    \includegraphics[scale=0.8]{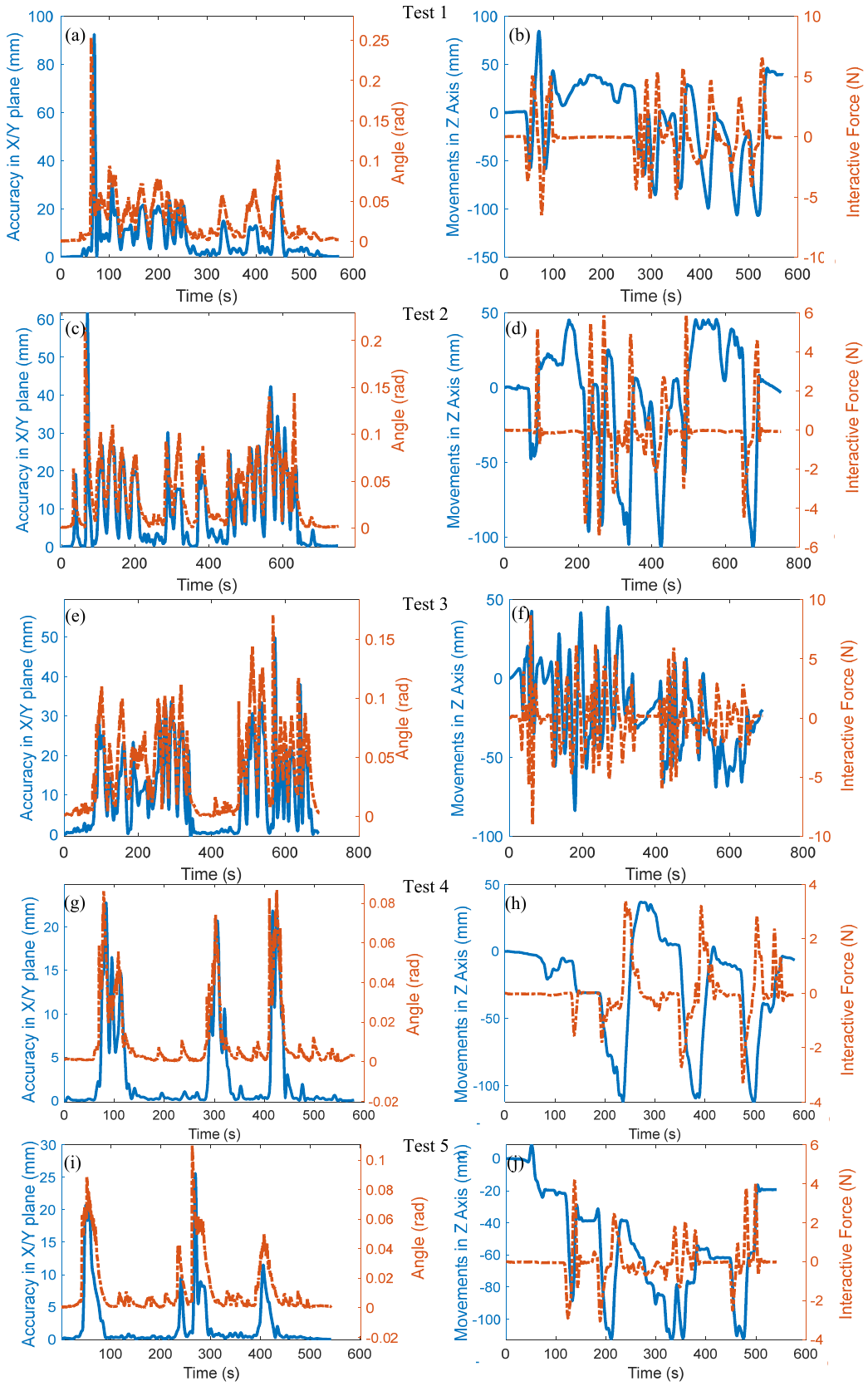}
    \vspace{-10pt}
    \caption{Results of the three sub-tasks during collaborative drilling. The left column figures show the robot's planar positioning accuracy (blue) and orientation accuracy (orange), while the right column figures illustrate the guiding force of the surgeon (orange) and the positional changes of the feeding direction executed by the robot (blue).}
    \label{tracking}
    \vspace{-10pt}
\end{figure*}

In terms of quantitative indicators in positions, orientations and force performance, the corresponding results were presented in Fig. \ref{tracking}. 
The force sensor collects the HRI force in real-time, while the differences between the position/orientation commanded to the robot and the actual position/orientation achieved by the robot are recorded as positional and orientation errors.

The vertical axis of the blue line in the Fig. \ref{tracking} (a),(c),(e), (g),(i) (i.e. left column) represents the planar distance from the origin to the target position projected onto the plane formed by the x/y axes of the surgical tool frame. This distance represents the error between the target and the surgical tool during the alignment process. The horizontal axis of the blue line on the left column of Fig.\ref{tracking} represents time, in seconds. In the experiment corresponding to the blue line on the left column of Fig.\ref{tracking}, the distance error fluctuates due to the intentional shaking of the skull phantom intraoperatively, ensuring the robotic manipulator is constantly aligned with it. The orange line on the left column of Fig.\ref{tracking} represents the same process as the blue line, but its vertical axis is angular error, which is reflected by the angle in the axis-angle representation of the difference between two frame orientation matrices shown in Fig.\ref{figExperiment}.(c). The vertical axis of the blue line in the Fig.\ref{tracking} (b),(d),(f),(h),(j) (i.e. right column) represents the projection of the robot's drilling motion trajectory onto the Z-axis of the surgical tool. The orange line on the right column of Fig.\ref{tracking} represents the change in force applied by the surgeon while dragging the robot to drill.

The maximum HRI force recorded was 9 N, with positional and orientation errors peaking at 1.06 mm and 0.0064 rad, respectively.
Notably, the maximum values for positional and orientation refer to the errors observed when the system was stable.

\subsection{Drilling Accuracy}
\begin{figure}[t]
	\centering
	\includegraphics[width=\columnwidth]{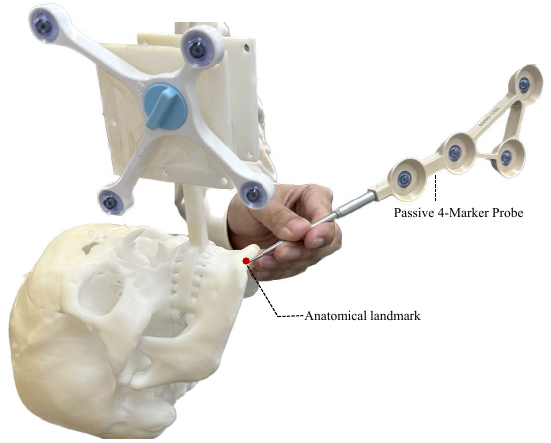}
     \vspace{-10pt}
	  \caption{Measurement of drilling accuracy. Multiple measurements were taken at the center point of the burr hole using a Passive 4-Marker Probe, and multiple measurements were also taken at the corresponding anatomical landmark on the patient's mandibular.}
	  \label{figdrilling}
    \vspace{-10pt}
\end{figure}

The first three types of experiments verified the precision of the robotic system at different stages of operation.
In this experiment, we utilized an anatomical landmark as a target point that is readily identifiable in both medical imaging and on a skull phantom.
The experimental workflow was as follows: Firstly, select an anatomical landmark on the patient's mandibular as the target point (drilling point). Subsequently, the pipeline of robotic drilling mentioned in Section \ref{sec:coopdrillexp} was executed. After drilling, a burr hole was left on the skull. As the burr hole and target point could be seen on the skull phantom surface. To evaluate the drilling accuracy, we use the Passive 4-Marker Probe of OTS to measure the distance from the burr hole to the target point, as shown in Fig. \ref{figdrilling}. The positional errors of 5 experiments were 1.56mm, 1.66 mm, 1.81 mm, 2.03 mm, 2.19mm. The average error was 1.85 mm while the maximum error was 2.19 mm which conforms to the accuracy requirements of MASO.

\section{Discussion}
\label{sec:discussion}

There are 2 aspects we considered in the proposed system: accuracy and compliance in collaborative operation. In this section, we discuss each in turn.

\subsection{Accuracy}
In terms of accuracy, the results showed 1.85 mm level (2.19 mm for maximum) overall drilling accuracy. The overall accuracy consists of patient registration errors, hand-eye calibration errors, and tracking errors, which are 0.32, 0.65, and 1.06 mm. 

The errors in patient registration may arise from two aspects: on one hand, since our method employs a ball-center finding regression approach, there is an inherent error in solving the center regression problem for irregular geometric objects (the lenses are not strictly semi-balls); on the other hand, the results of manual segmentation cannot perfectly match the true shape of the object, resulting in model error.

Hand-eye calibration is critical for surgical robotic planning and control. While global space performance may not be flawless, the accuracy within the robot's operational workspace can be improved by strategically sampling points around the workspace. This includes uniformly selecting points along the workspace boundaries and at its center.
In our experiments, we recorded five sampled points per calibration and collected additional five testing points to verify accuracy. The number of sampled points gives a balance between calibration accuracy and time efficiency. Increasing the number of points enhances accuracy but at the cost of extended preparation time. Conversely, reducing the number of points speeds up the registration process but may compromise accuracy. We determined our optimal number of sampled points, $N = 5$, through empirical evidence and experimental validation.
Calibration errors stem from inherent sensor measurement inaccuracies (in our case, the OTS) and tracker installation errors, as well as the configuration of the design matrix in the regression task in Equation \eqref{equ7b}.
To optimize the design matrix, the position and orientation of trackers should be uniformly distributed, which aids in avoiding singularity and ensuring numerical stability for the regression task.

During the tracking procedure, errors primarily originated from the robotic controller. Disturbances caused by the movement of skull clamp and crosstalk between different subtasks did not significantly impact accuracy, particularly when the robot reached a steady state (a period characterized by nearly horizontal positioning). 
Owing to the clear delineation of positional control, orientation control and force control tasks, the robot was able to concurrently execute these tasks in their respective subspaces without compromising on precision, maintaining steady errors around 1mm.

Additionally, the frame rate of OTS is 30Hz, and in the case of using dynamic measurement mode, there is a certain delay error, resulting in some disturbances in the information transmitted to the robot controller.
Since the robot requires some time to reach a stable state when positioning, drilling immediately after positioning during the experiment could result in errors due to the robot's unstable state. This can be seen in the data on the left column of Fig. \ref{tracking}, the nearly horizontal lines represent the robot reaching a stable state. Based on this analysis, the experiment's errors will fluctuate depending on the skull movement.
The movement of the skull during operation will lead to the robotic tracking, and the relative position between robot end effector and the desired drilling point will converge to zero as the robot reaches a stable state.



\subsection{Compliance in Collaborative Operation}
Controlled by the surgeon, the feeding velocity introduces an element of intraoperative supervision.
This method is both intuitive and expedient. 
Surgeons can manually adjust the robotic feeding or retreating by the handle, initiating or halting tasks in a manner akin to using traditional surgical tools. 
Additionally, surgeons receive haptic feedback upon the robot's contact with bone, aiding in the management of the drilling process and enhancing their awareness of the task's progression. Furthermore, the robot alleviates physical strain by compensating for tool weight, thus minimizing errors linked to surgeon fatigue and hand tremors.


\section{Conclusion}
\label{sec:conclusion}
A novel robot-assisted system for MASO has been developed. To help surgeons finish the drilling task during the MASO surgeries, collaborative operation has been applied in this system. The collaborative operation can be decomposed into three sub-tasks: positional control and orientation control, both led by the robot for precise alignment; and force control, managed by surgeon to preserve the tactile feedback of the drilling operation as much as possible, ensures a prompt response to unexpected emergencies. At the same time, the tool's weight is borne by the robot, to reduce the physical burden on the surgeon and eliminate the impact of hand tremors on surgical precision. Another key point of this paper is the introduction of the occlusal splint, which allows for a more reasonable fixation of the optical-based tracker to the patient, and eliminates the use of stereotactic positioning devices in MASO, thereby reducing patient trauma.
Experiments were performed to evaluate the system from various aspects. The maximum overall error of the patient registration was 0.32mm, while the hand-eye calibration error arrived at 0.65mm. The tracking experiments showed the robot performance on positioning and the change of guiding force.
Phantom experiments revealed that the robot-assisted system could conduct the drilling operation with high accuracy, with 1.85 mm operation accuracy, which meets clinical needs. 

Although the system proposed in this paper shows clear advantages in terms of positioning accuracy, intuitive drilling, and reduced operational effort in MASO surgeries, its application still has some limitations. 
For tasks that require operation in narrow spaces, the optical tracker is bulky. To enable the optical tracker to be used in these scenarios, additional efforts are often required, such as designing extension parts to ensure the tracker has sufficient space for installation. However, these intermediate connectors can lead to a decrease in accuracy. 
Additionally, the optical tracker used in this paper is equipped with 4 Radix Lens, and the system remains unaffected if any one of them is obstructed. However, if two or more Radix Lens are obstructed, the optical tracking system will fail to supply the pose estimation of tracker, which can significantly impact the surgeon's operation.


In future research, to further expand the operational scope of the robot, considerations should be made towards miniaturizing markers. This is particularly effective in tasks involving operations in confined spaces. On the other hand, the exploration of markerless technology can be conducted, utilizing the anatomical structural features of the human, such as dental, to eliminate the constraints of external markers. Such advancements are promising to enhance the system’s adaptability and precision in more complex surgical environments.
Besides, the application of this system can be expanded to tasks similar to the drilling operations during MASO surgeries, such as dental implant surgeries after modifications to the surgical instrument's end connection. It can also be applied in spinal surgeries for pedicle screws placement, where, after locating the screw channels, surgeons can utilize a Kirschner drill for drilling; in the preparatory tasks of craniotomy surgeries, where surgeons need to open a bone window, this system can also accomplish the bone drilling and window cutting task after replacing the corresponding surgical tools.


\section*{APPENDIX}
The detailed derivation process from Equations \eqref{equ1a} and \eqref{equ1b} to Equations \eqref{equ2a} and \eqref{equ2b} is as follows:
\begin{equation}    
\begin{aligned}
H(\mathbf{x_c}, R) &=\sum_{i=1}^{N_{S}}(||\mathbf{x}_i-\mathbf{x_c}||_{2}^{2} - R^{2})^{2} \\
&=\sum_{i=1}^{N_{S}}((\mathbf{x}_i-\mathbf{x_c})^{T}(\mathbf{x}_i-\mathbf{x_c}) - R^{2})^{2} \\
&=\sum_{i=1}^{N_{S}}(\mathbf{x}_i^{T}\mathbf{x}_i-2\mathbf{x}_i^{T}\mathbf{x}_c+\mathbf{x}_c^{T}\mathbf{x}_c-R^{2})^{2} \\
\label{equaa}
\end{aligned}
\end{equation}
when the $\frac{\partial H}{\partial \mathbf{x_c}} = \mathbf{0}$, the Equation \eqref{equaa} could be simplified as:
\begin{equation}    
\begin{aligned}
\sum_{i=1}^{N_{S}}2(\mathbf{x}_i^{T}\mathbf{x}_i-2\mathbf{x}_i^{T}\mathbf{x}_c+\mathbf{x}_c^{T}\mathbf{x}_c-R^{2})(-2\mathbf{x}_i+2\mathbf{x}_c) =0 
\label{equbb}
\end{aligned}
\end{equation}
when the $\frac{\partial H}{\partial R} = \mathbf{0}$, the Equation \eqref{equaa} could be simplified as:
\begin{equation}    
\begin{aligned}
\sum_{i=1}^{N_{S}}2(\mathbf{x}_i^{T}\mathbf{x}_i-2\mathbf{x}_i^{T}\mathbf{x}_c+\mathbf{x}_c^{T}\mathbf{x}_c-R^{2})(-2\mathbf{R}) =0 
\label{equcc}
\end{aligned}
\end{equation}
Due to $-2\mathbf{R} \neq 0$, the Equation \eqref{equcc} could be expressed as follows:
\begin{equation}    
\begin{aligned}
\sum_{i=1}^{N_{S}}\mathbf{R}^{2} &=\sum_{i=1}^{N_{S}}\mathbf{x}_i^{T}\mathbf{x}_i-(2\sum_{i=1}^{N_{S}}\mathbf{x}_i^{T})\mathbf{x}_c+\sum_{i=1}^{N_{S}}\mathbf{x}_c^{T}\mathbf{x}_c \\
{N_{S}}\mathbf{R}^{2}&=\sum_{i=1}^{N_{S}}\mathbf{x}_i^{T}\mathbf{x}_i-(2\sum_{i=1}^{N_{S}}\mathbf{x}_i^{T})\mathbf{x}_c+{N_{S}}\mathbf{x}_c^{T}\mathbf{x}_c \\
\mathbf{R}^{2} &=\frac{1}{N_s}\sum_{i=1}^{N_{S}}\mathbf{x}_i^{T}\mathbf{x}_i- (\frac{2}{N_s}\sum_{i=1}^{N_{S}}\mathbf{x}_i^{T})\mathbf{x}_c+\mathbf{x}_c^{T}\mathbf{x}_c
\label{equdd}
\end{aligned}
\end{equation}
Use $M_1$ to represent $\frac{1}{N_s}\sum_{i=1}^{N_{S}}\mathbf{x}_i^{T}\mathbf{x}_i$, and $M_2$ to represent $\frac{1}{N_s}\sum_{i=1}^{N_{S}}\mathbf{x}_i^{T}$. Substitute Equation \eqref{equdd} into Equation \eqref{equbb} to eliminate the $\mathbf{R}^{2}$, resulting in Equation \eqref{equee}.
\begin{equation}    
\begin{aligned}
\sum_{i=1}^{N_{S}}(\mathbf{x}_i^{T}\mathbf{x}_i\mathbf{x}_i-2\mathbf{x}_i^{T}\mathbf{x}_c\mathbf{x}_i-\mathbf{M_1}\mathbf{x}_i+2\mathbf{M_2}\mathbf{x}_c\mathbf{x}_i)=0 
\label{equee}
\end{aligned}
\end{equation}
To obtain the coordinate of $\mathbf{x_c}$, the Equation \eqref{equee} could be transformed as follow:
\begin{equation}    
\begin{aligned}
\sum_{i=1}^{N_{S}}2\mathbf{x}_i^{T}\mathbf{x}_c\mathbf{x}_i-\sum_{i=1}^{N_{S}}2\mathbf{M_2}\mathbf{x}_c\mathbf{x}_i = \sum_{i=1}^{N_{S}}(\mathbf{x}_i^{T}\mathbf{x}_i\mathbf{x}_i-\mathbf{M_1}\mathbf{x}_i)
\label{equff}
\end{aligned}
\end{equation}
The left side of Equation \eqref{equff} can be simplified to Equation \eqref{equ2a}, while the right side of Equation \eqref{equff} can be simplified to Equation \eqref{equ2b}.

\section*{Acknowledgments}
This work was supported in part by National Key R\&D Program of China under Grant 2022YFC240014102, in part by the National Natural Science Foundation of China under Grant 62073043, in part by Beijing Institute of Technology under Chasing Dream Abroad Scholarship.




\bibliographystyle{cas-model2-names}

\bibliography{refs.bib}

\end{document}